*Original Article*

# Improved Forecasting Using a PSO-RDV Framework to Enhance Artificial Neural Network

Sales G. Aribe Jr.

*Information Technology Department, Bukidnon State University, Malaybalay City, Bukidnon, Philippines.*

*Corresponding Author : sg.aribe@buksu.edu.ph*

*Abstract* - Decision-making and planning have long relied heavily on AI-driven forecasts. The government and the general public are working to minimize the risks while maximizing benefits in the face of potential future public health uncertainties. This study used an improved method of forecasting utilizing the Random Descending Velocity Inertia Weight (RDV IW) technique to improve the convergence of Particle Swarm Optimization (PSO) and the accuracy of Artificial Neural Network (ANN). The IW technique, inspired by the motions of a golf ball, modified the particles' velocities as they approached the solution point to a parabolically descending structure. Simulation results revealed that the proposed forecasting model with [0.4, 0.9] combination of alpha and alpha_dump exhibits a 6.36% improvement in position error and 11.75% improvement in computational time compared to the old model, thus improving its convergence. It reached the optimum level at minimal steps with a 12.50% improvement as against the old model since it provides better velocity averages when speed stabilization occurs at the 24th iteration. Meanwhile, the computed p-values for NRMSE (0.04889174), MAE (0.02829063), MAPE (0.02226053), WAPE (0.01701545), and $R^2$ (0.00000021) of the proposed algorithm are less than the set 0.05 level of significance; thus the values indicated a significant result in terms of accuracy performance. Applying the modified ANN-PSO using the RDV IW technique greatly improved the new HIV/AIDS forecasting model compared with the two models.

*Keywords* - Artificial Neural Network, Particle Swarm Optimization, Inertia Weight, forecasting, HIV.

## 1. Introduction

With a 96.56% increase in HIV incidence between 2009 and 2022, the Philippines is dealing with the quickest-growing HIV epidemic in the Western Pacific [1, 2]. Since then, 109,282 total reported cases have been recorded by the HARP as of December 2022 [3]. HIV-related deaths, new HIV infections, and the cost of antiretroviral therapies (ART) have all considerably decreased as a result of efforts to end HIV transmission. Unfortunately, the HIV epidemic in the country continues to be shockingly severe due to the dearth of HIV preventive programs and diagnostic and counseling services. Due to these problems, it is difficult to encourage patients to undergo screenings and maintain high levels of ART compliance. Additionally, COVID-19 restrictions, lack of information, and pervasive stigma are blamed for making it difficult to get treatment. Modeling the course of HIV/AIDS and predicting its pace of spread have a significant impact on health systems and lawmakers. The formulation of policies depends on the insights provided by forecasting systems, guiding the creation of novel public health measures and strategies while also assessing the effectiveness of implemented policies [4]. Knowing the probable course of the spread will aid governments and health professionals in coming up with potential treatments and precautionary measures. The timely communication of crucial events, such as HIV forecasts and newly implemented policies, is paramount for effective dissemination.

With various ML techniques available, the ANN proved its superiority in terms of higher acceptable values for accuracy in forecasting [5, 6]. Due to interest in using ANN for forecasting outcomes, the amount of research conducted has significantly increased in recent years. The ANN-based model is an effective technique with greater precision [7]. It would be the most widely used and standard network model in numerous applications, such as stock price prediction [8, 9], oil price [10], air quality index [11, 12], COVID-19 spread [13], HIV/AIDS prediction and many others. However, despite many of the benefits mentioned, some real-time series do not perform satisfactorily with ANN [14].

Forecasters have a crucial but frequently challenging duty of enhancing forecasting accuracy, particularly for time series. Unfortunately, ANN's primary drawback is that it requires an enormous amount of time and exclusively relies on learning through trial and error [15], therefore resulting in poor predictive accuracy performance. Typically, its parameters are insufficient to produce accurate models, necessitating the replacement of the ANN's architecture with one that accelerates each particle to its personal and global optimal locations at each subsequent step.

Any optimization method can be employed to update the architectures and weights of ANN [16]. Therefore, in the previous relevant literature, some Swarm Intelligsence (SI) approaches were utilized by ANN to enhance the parameters because it has a significant effect on its performance.

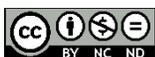

To address this problem, several optimization methods were employed to improve the forecasting output's accuracy of the ANN model. These optimizers include the Whale Optimization Algorithm (WOA) [17], Multi-objective (MO) [18], Modified Genetic Algorithm (MGA)[19], Firefly Algorithm (FA) [9], Imperialism Competitive Algorithm (ICA) [20], Multi-Objective Grey Wolf Optimizer (MOGWO) [21], and many others. However, since these optimization techniques need an extensive number of additional parameters and iterations, feature scaling can affect how they perform [22]. Moreover, these algorithms also suffer from computational complexity. Among these optimizers, PSO proved to be superior in terms of the following: straightforward concept, simple implementation, robustness to control variables, and speed of computation in comparison with other metaheuristic (MH) optimization methods and mathematical algorithms [23].

To enhance the accuracy of the ANN model, researchers have turned to SI approaches with PSO as a prominent optimizer. It is utilized in a number of predicting fields, such as financial forecasting [24], prediction of gas density [25], overbreak prediction in underground excavations [26], biomass higher learning value prediction [27], and air quality forecasting [28], among others. PSO has finally been a popular choice among academics and has evolved to offer good performance in a variety of application domains. It also has the capacity to hybridize, specialize, and exhibit some endearing emergent behaviors. With PSO's efficient use of memory and easy implementation, it is effective in achieving the global minimum with fewer iterations [29]. Although PSO stands out for having few parameters, performing quickly, and generating results that are effective [30], it also endures from the following limitations. It converges very slowly to the universal solution point and is unable to find the optimal global solution [31], which results in poor convergence [32].

Another problem that degrades the performance of PSO is its local optima entrapment because the strengths of exploration and exploitation are not properly balanced [33]. In order to advance towards the best solution, exploration needs to conduct a thorough search in the early stages of the search process [34]. PSO algorithms must possess robust exploration capabilities in order to prevent becoming stuck in a neighboring local optimal. Exploitation, on the other hand, focuses on areas which have an excellent probability of being the target point where the optimum solution can be located. Finding the best answer entails finding the right balance between exploitation and exploration [35]. This problem is due to the absence of a parameter that regulates the movement of the particle velocity [36]. By incorporating new IW parameters like alpha, alpha_dump, and delta that are driven by the motions of the golf ball, the velocity has been enhanced to address this shortcoming [29, 37]. This will control the influence of a particle's previous momentum on its future trajectory [38].

Thus, it will improve its search capacity (exploration) while concentrating on regions with the optimal solution (exploitation).

To summarize, ANN suffers from poor accuracy performance due to the absence of parameters that regulate the weights and accelerate it towards its best local and global position. Optimizing it using PSO offers additional parameters to obtain good performance; however, PSO also endures from poor convergence due to a lack of parameters that regulate the particle velocity and converge it at the solution point in a few iterations. Thus, this study applies the meta-heuristic optimization technique called RDV IW technique to overcome these limitations.

## 2. Materials and Methods
### 2.1. Data

The first two instances of HIV incidence in the Philippines occurred in 1984, while ART treatment only started in 2009 [39]. In order to forecast the effect of ART coverage in the Philippines, the data spanning March 2009 to December 2022 (166 months) was utilized for training and testing.

In this study, HIV dataset were obtained from the official monthly report released by HARP which has official records of each laboratory-confirmed diagnosis, all HIV-related deaths, and the status of ART in the Philippines. Confirmed HIV-positive persons were documented in HARP and submitted to the Philippines DOH Epidemiology Bureau. Records from 180 treatment facilities and primary HIV care treatment institutions that submitted HARP reports during the reporting period contain counts of HIV patients who are currently enrolled and receiving antiretroviral (ARV) medication during that time.

### 2.2 Artificial Neural Network

This study utilizes an ANN forecasting model consisting of three components: a summing element, an activation function or f(), and a network of synapses facilitating the passage of input connected to the neuron and multiplied by its weight, $w_i$, $x_i$.

The set of weights, denoted as $w_1, w_2...w_i$, corresponds to connection strengths, while the set of inputs is represented by $x_1, x_2...x_i$. A high value is assigned to the weight when the input strength is robust, and conversely, a lower value is assigned when the input strength is weak. The inputs are therefore multiplied by weights and represented by Equations 1 to 4:

$$S = w_1 x_1 + w_2 x_2 + \cdots \text{next}, = \sum w_i x_i \quad (1)$$

$$S = \sum_{i=1}^{n}(w_i x_i - \theta) \quad (2)$$

$$S = f(x) = x \quad (3)$$

$$\delta = b - s = b - x \quad (4)$$

The threshold value is determined by the summation value in the equation above. With a single input and output, a neuron, equipped with an appropriate threshold value and weight, generates an output within a time unit. The threshold function is required in order to evaluate the variations of input and output. If the summation of the output, denoted as S(f(x)), exceeds the specified threshold, the output will be 1; otherwise, it will be 0. It can also be expressed as follows

$$f(x) = 1, x > 0 \quad (5)$$

$$f(x) = 0, x \leq 0 \quad (6)$$

This type of dynamic filtering involves predicting the future outcomes of any number of time series using their historical values. For nonlinear filtering and forecasting, neural networks employ tapped delay lines.

In the study, Python 3.11.1 is used to solve time series issues using the nonlinear autoregressive (NAR) tool. It forecasts the series time value of y(t) given the past value of y(t). Expressed mathematically, it can be represented as

$$yt = f(y(t-1) + y(t-2) + y(t-3) + y(t-m)) \quad (7)$$

An algorithm for estimating monthly instances of the spread of HIV was trained and tested using data spanning 166 months. The MLP was used, but the fundamental issue with using it is either underfitting or overfitting. The layers that provide input and output are inconsistent as a result of this. The Stop Training Approach (STA) was successfully used to minimize errors. Monthly HIV case information is gathered and separated into testing, validation, and training data at random.

### 2.3. Enhanced ANN Using Particle Swarm Optimization

In the PSO method, the search agents begin by randomly diversifying the particle locations ($x_i$) and velocities ($v_i$) around the search space in dimensions *j-th*. The particles then adjust their placements in accordance with the personal and global best positions as far discovered, as stated below.

$$x_{ij}^{(t+1)} = x_{ij}^{(t)} + v_{ij}^{(t+1)} \quad (8)$$

$$v_{ij}^{(t+1)} = v_{ij}^{(t)} + c_1 r_1 (x_{ij}^{p(t)} - x_{ij}^{(t)}) + c_2 r_2 (x_j^{g(t)} - x_{ij}^{(t)}) \quad (9)$$

In Equations 8 and 9, $x_{ij}$ and $v_{ij}$, respectively, stand for the location and motion of the *i-th* particle along the *j-th* dimension. The letter *t* stand for the present iteration, and the signs for acceleration coefficients $c_1$ and $c_2$ represent the individual and global weights. The ideal position of the agent (i) at dimension (j) is represented by the pbest, or $x_{ij}^{p(t)}$, while the best agent globally at dimension *j* as of yet is represented by the gbest, or $x_j^{g(t)}$. Random numbers $r_1$ and $r_2$ are selected within the range 0 to 1.

Accurate definition of the fitness function is crucial for the application of MH algorithms. The fitness function, denoted as $f_i$, serves to assess the distance between each particle and the optimal solution in problem-solving using these methods. "x" is the position (Equation 8), and "v" is the present particle velocity (Equation 9). These two crucial criteria determine whether the particle will converge. One factor that directly affects the final result of the solution is the overall count of particles. Because the result improves as the particle number increases, the algorithm's performance suffers.

### 2.4. Enhanced ANN-PSO Using Random Descending Velocity – Inertia Weight Technique

Enabling new parameters is necessary for PSO improvement. This study enhanced the velocity parameter by extracting novel parameters motivated by the ball movements throughout a round of golf. The ANN-PSO method is combined with the RDV technique in Pseudocode 1, where lines 11 to 15 are added. To control the particles, much like how the golf ball is controlled from its start movement until the game is over, the delta parameter in line 11 is calculated. As demonstrated in (Equation 10), it begins with a maximum value of 1 at the initiation of the algorithm but gradually diminishes over time, ultimately converging to a value approximately close to zero.

**Pseudocode 1: ANN-PSO-RDV IW Algorithm**

1. Read dataset
2. Partition the dataset into training and testing sets
3. Set up the initial values of $X_i$, $V_i$, iteration, pbest, gbest
4. Create random particles (P)
5. For every particle (i)
6.     Compute fitness function ($f_i$)
7.     Update the values of pbest, gbest
8. End for
9. While iteration
10.     For each particle i
11.         *Calculate delta*
12.         *If delta < rand, then alpha = alpha * alpha_dump*
13.         *W = alpha*
14.         *Update $V_i = V_i * W$*
15.         *Update $X_i$*
16.         If $X_i$ > limit, then $X_i$ = limit
17.         // ANN
18.         Create ANN model
19.         Train the model
20.         Test the model
21.         Calculate fitness function $f_i$
22.         Update pbest, gbest using position error (PE) as fitness value
23.     End for
24. End while
25. Inverse transform predictions
26. Compute evaluation metrics

The delta parameter serves to represent the movement of the golf ball in a game as it approaches the target location. Particles, for instance, roll around at typical speeds at the beginning of an iteration, similar to how a golf ball does on impact. However, owing to randomness, the value of delta will be compared with a randomly generated value.

Given that delta reaches its maximum value at the beginning of the iteration, it is expected to exceed the randomly chosen value within the range of 0 to 1. During the initial iterations of this process, the particles are allowed to move at moderate speeds.

The randomly generated value becomes increasingly noticeable as the iteration draws to a close since the delta has the lowest value. By employing the "*alpha_dump*" parameter, particle velocities are thereby somewhat reduced while minimizing the extent of alterations in particle positions.

$$\Delta = \frac{-\text{iteration}}{e^{\max_{\text{iteration}}}} \quad (10)$$

The alpha parameter utilized in this study was chosen at random from [0, 1] and assigned a value of 0.4, whereas the alpha_dump parameter was chosen at random from [0.5, 1] and assigned a value of 0.9. This parameter is employed to intermittently decrease the value of alpha, representing speed, by a specified extent.

If the delta value surpasses the randomly drawn number, as depicted in Equation 11, the alpha can be reduced, as illustrated in line 11 of Pseudocode 1. The goal of all of these calculations is to slow the particle down. As a result, Equation 9 is amended to include the IW parameter represented by *w*, as illustrated in Equation 12.

$$\alpha = \alpha * \text{alpha\_dump} \quad (11)$$

$$v_{ij}^{(t+1)} = wv_{ij}^{(t)} + c_1 r_1 (x_{ij}^{p(t)} - x_{ij}^{(t)}) + c_2 r_2 (x_j^{g(t)} - x_{ij}^{(t)}) \quad (12)$$

### 2.5. Evaluation Metrics

This study uses 2 types of performance evaluation metrics: accuracy and convergence. The accuracy of the proposed model is determined using Equation 13 to 17, and its convergence rate is measured using Equation 18. Computational time and number of iterations as an attribute of good convergence are performed and produced automatically by Python.

$$\text{NRMSE} = \sqrt{\frac{\sum (S_i - O_i)^2}{\sum O_i^2}} \quad (13)$$

$$\text{MAE} = \frac{1}{N} \sum_{i=1}^{N_s} |P_{yi} - Y_i| \quad (14)$$

$$\text{MAPE} = \frac{1}{N} \sum_{i=1}^{N} \left|\frac{P_{yi} - Y_i}{YP_i}\right| \quad (15)$$

$$\text{WAPE} = \frac{\sum_{i=1}^{N} |y_i - p_i|}{\sum_{i=1}^{N} |y_i|} \quad (16)$$

$$R^2 = 1 - \frac{\sum_{i=1}^{n}(Y_i - P_{yi})^2}{\sum_{i=1}^{n}(Y_i - \overline{Y}_i)^2} \quad (17)$$

$$\text{PE} = \sqrt{(x^2 - x^1) + (y^2 - y^1) + (z^2 - z^1)} \quad (18)$$

## 3. Results and Discussion

ANN, PSO, and RDV IW models were applied to the series to create a reliable forecasting model for the prevalence of HIV in the Philippines. Finding the optimal combination required a careful examination of parameter estimation. This section also describes how the suggested model performed in terms of accuracy and convergence, employing several assessment criteria.

**Table 1. Position error of alpha and alpha_dump combinations**

| α   | α_dump | Position Error | α   | α_dump | Position Error |
|-----|--------|----------------|-----|--------|----------------|
| 0.1 | 0.5    | 104.581589     | 0.6 | 0.5    | 106.844099     |
| 0.1 | 0.55   | 112.226553     | 0.6 | 0.55   | 103.632794     |
| 0.1 | 0.6    | 106.798350     | 0.6 | 0.6    | 103.885669     |
| 0.1 | 0.65   | 108.64161      | 0.6 | 0.65   | 106.755019     |
| 0.1 | 0.7    | 114.446026     | 0.6 | 0.7    | 106.869555     |
| 0.3 | 0.5    | 115.012677     | 0.8 | 0.5    | 112.712593     |
| 0.3 | 0.55   | 112.693036     | 0.8 | 0.55   | 109.426907     |
| 0.3 | 0.6    | 110.046214     | 0.8 | 0.6    | 108.886974     |
| 0.3 | 0.65   | 107.757651     | 0.8 | 0.65   | 109.166934     |
| 0.3 | 0.95   | 116.854028     | 0.8 | 0.95   | 105.617997     |
| 0.4 | 0.5    | 113.982131     | 0.9 | 0.5    | 115.285844     |
| 0.4 | 0.55   | 110.453716     | 0.9 | 0.55   | 114.545124     |
| 0.4 | 0.6    | 109.269132     | 0.9 | 0.6    | 111.997822     |
| 0.4 | 0.65   | 108.815269     | 0.9 | 0.65   | 108.652044     |
| 0.4 | 0.7    | 112.375528     | 0.9 | 0.7    | 103.773858     |
| 0.4 | 0.9    | 101.273250     | 0.9 | 0.9    | 106.190214     |
| 0.4 | 0.95   | 109.335953     | 0.9 | 0.95   | 112.910467     |
| 0.5 | 0.75   | 110.188497     | 1.0 | 0.75   | 107.654552     |
| 0.5 | 0.8    | 112.619053     | 1.0 | 0.8    | 113.069394     |
| 0.5 | 0.85   | 112.375092     | 1.0 | 0.85   | 105.27955      |
| 0.5 | 0.9    | 108.815649     | 1.0 | 0.9    | 124.924420     |
| 0.5 | 0.95   | 106.603585     | 1.0 | 0.95   | 112.065274     |

## 3.1. RDV IW Optimal Parameter Combination

In the research conducted by Dereli and Köker, the values of alpha = 0.3 and alpha_dump = 0.95 produced the best results. These parameter values worked well together to somewhat reduce the velocity value of the particle in the PSO algorithm. The value of position error as the measure of convergence performance was used as the deciding factor for choosing the optimal combination. Based on the preliminary experiment, the combination did not provide good forecasting output, as evident in the poor values for accuracy and convergence evaluation metrics. This prompted the researcher to explore with alternative pairs.

Table 1 shows 100 combinations of RDV IW control parameters in terms of alpha between [0.1, 1.0] and alpha_dump between [0.5, 1.0] and their corresponding position error. In this table, [alpha, alpha_dump] combinations are highlighted in bold and bigger font sizes. The default parameter values [0.3, 0.95] produced a suboptimal combination, as evidenced by its extremely high position error.

Surprisingly, it was the fourth-worst pair out of 100 combinations. On the other hand, the worst combination was [1.0, 0.9], which generated the highest position error. Thus, the optimal combination used in this study was [0.4, 0.9], which obtained the least position error.

## 3.2. Convergence Performance

In this study, the position error, number iterations, and computational time of the random descending velocity PSO approach are thoroughly investigated. In order to address this, the test procedures have included the comparison of results with traditional PSO alongside the results of this technique. Experimental results for convergence in terms of Position Error are shown in Table 2. The min, max, and average were noted and compared after performing 10 independent runs. Results from the algorithm achieving the optimal performance metrics are presented in bold and larger text sizes.

As shown in the table, the minimum, maximum, and average values produced by this study were better than those of hybrid ANN-PSO after 10 runs. As such, the proposed algorithm in all conditions was found to be closer to the target points with respect to Euclidean distance. Specifically, it obtains the least position error, equivalent to 101.2732506, compared to the traditional PSO, equivalent to 107.71067, with a 6.36% improvement.

This can be supported by Figure 1, which shows and compares the convergence plot between two models, where the blue line corresponds to the ANN-PSO model, while the red line corresponds to the ANN-PSO implemented with the RDV IW technique.

**Table 2. Comparative analysis of this study and ANN-PSO as to position error**

| Condition | Position Error | | Improvement |
|---|---|---|---|
| | *This Study* | *ANN-PSO* | |
| Min | 101.2732506 | 107.71067 | 6.36% |
| Max | 111.46789 | 120.07746 | 7.72% |
| Average | 108.99974 | 110.93828 | 1.78% |

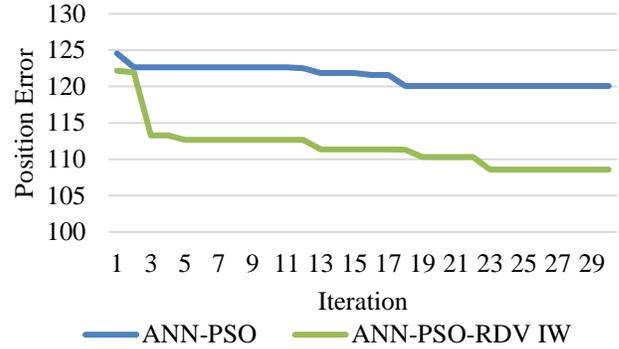

Fig. 1 Convergence plot for the position error

Figure 1 illustrates that PSO with RDW IW consistently yields superior results compared to PSO without IW, irrespective of the number of iterations. Although the distinction between the two models is minimal after the initial iteration, it progressively widens with subsequent iterations. Therefore, it is reasonable to conclude that the RDV IW technique significantly enhances the performance of PSO.

For convergence with respect to calculation time this is shown in Table 3. The min, max, and average were also noted and compared after performing 10 independent runs. Table 3 unequivocally demonstrates how excellent the proposed algorithm is, not only in obtaining the least position error but also in computational time (in seconds) in all aspects of conditions vis á vis the number of iterations.

With respect to a minimum, this study obtains the shortest computational time of 210.01812 seconds compared to 234.68989 seconds of ANN-PSO, an 11.75% improvement. Figure 2, which depicts the computation periods graph during which the IW strategies put to the test also managed to obtain the least position error, which can help to corroborate this claim further. This figure further demonstrates how the RDV IW technique is superior in terms of both calculation time and position error.

**Table 3. Comparative analysis of this study and ANN-PSO as to computational time**

| Condition | Computational Time (in seconds) | | Improvement |
|---|---|---|---|
| | *This Study* | *ANN-PSO* | |
| Min | 210.01812 | 234.68989 | 11.75% |
| Max | 1,678.90319 | 2,169.69820 | 29.23% |
| Average | 511.49391 | 827.92385 | 61.86% |

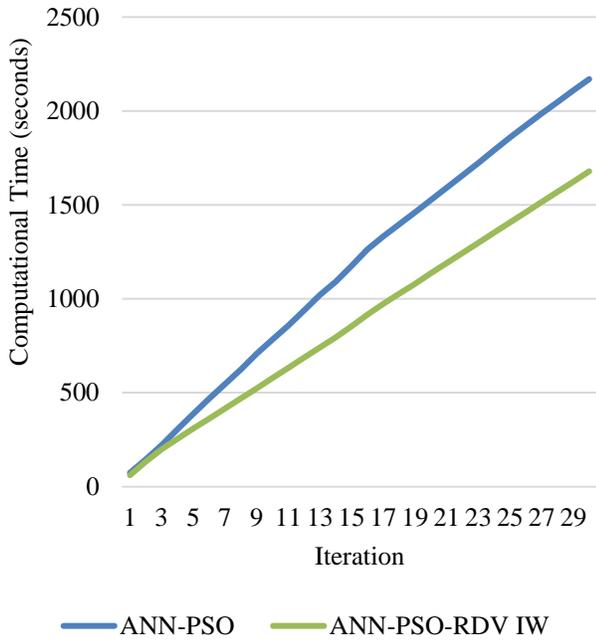

**Fig. 2 Convergence plot for the computational time**

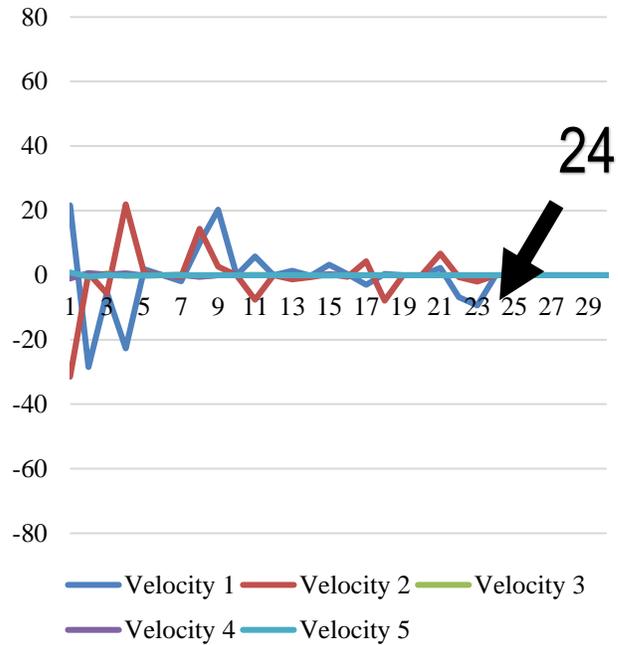

**Fig. 3 Velocity averages of PSO with RDV IW technique**

The novel approach put forward in this work is centered on minimizing variations in particle velocity values at locations close to the ideal value. This is a desirable characteristic for successful convergence. This reasoning serves as the foundation for this golfing trick for placing the ball in the target hole. The generated random numbers will probably be less than Delta because Delta is at its highest value in the initial iterations, just like the opening swing in the golf game.

In this situation, the particles will approach the target at their regular rates. As the delta approaches its minimal values at the end of the iteration, the randomly generated number will be greater than the delta.

In this scenario, the particle velocities will drop, and they will arrive at the desired place (PE) in the fewest possible steps (computational time and iterations). The least PE and shortest computational time are the key performance indicators of a faster convergence rate. For ANN-PSO applied with RDV IW technique, it can be depicted in Figure 3 how the proposed method was able to reach the stabilization stage at minimum steps. The suggested algorithm's speed averages have stabilized in its 24th iteration, compared to the classical PSO's 27th iteration, a 12.50% improvement. As a result, it appears that the position error value achieved is significantly superior to the previous model. Therefore, it was able to reach or converge at the target point with few iterations or minimal steps.

**Table 4. Accuracy scores of the proposed algorithm vs. ANN-PSO model**

| Trial | This Study | | | | | ANN-PSO | | | | |
|---|---|---|---|---|---|---|---|---|---|---|
| | *NRMSE* | *MAE* | *MAPE* | *WAPE* | $R^2$ | *NRMSE* | *MAE* | *MAPE* | *WAPE* | $R^2$ |
| 1 | 0.192 | 77.944 | 0.085 | 0.082 | 0.784 | 0.197 | 78.576 | 0.085 | 0.083 | 0.653 |
| 2 | 0.191 | 79.662 | 0.085 | 0.084 | 0.789 | 0.197 | 85.502 | 0.090 | 0.090 | 0.652 |
| 3 | 0.199 | 82.322 | 0.089 | 0.087 | 0.738 | 0.197 | 83.130 | 0.089 | 0.087 | 0.654 |
| 4 | 0.196 | 80.584 | 0.086 | 0.083 | 0.757 | 0.197 | 80.665 | 0.087 | 0.085 | 0.650 |
| 5 | 0.195 | 81.645 | 0.087 | 0.086 | 0.765 | 0.200 | 81.913 | 0.088 | 0.086 | 0.633 |
| 6 | 0.194 | 81.832 | 0.089 | 0.086 | 0.772 | 0.198 | 83.397 | 0.091 | 0.088 | 0.645 |
| 7 | 0.200 | 81.734 | 0.086 | 0.086 | 0.730 | 0.216 | 90.368 | 0.096 | 0.095 | 0.623 |
| 8 | 0.198 | 81.164 | 0.084 | 0.085 | 0.744 | 0.193 | 81.507 | 0.087 | 0.086 | 0.675 |
| 9 | 0.198 | 82.158 | 0.088 | 0.087 | 0.746 | 0.195 | 82.396 | 0.089 | 0.087 | 0.662 |
| 10 | 0.194 | 81.909 | 0.088 | 0.086 | 0.771 | 0.203 | 83.457 | 0.089 | 0.088 | 0.615 |
| **Min** | **0.191** | **77.944** | **0.084** | **0.082** | **0.730** | 0.193 | 78.576 | 0.085 | 0.083 | 0.615 |
| **Max** | **0.200** | **82.322** | **0.089** | **0.087** | **0.789** | 0.216 | 90.368 | 0.096 | 0.095 | 0.675 |
| **Average** | **0.196** | **81.096** | **0.087** | **0.085** | **0.760** | 0.199 | 83.091 | 0.089 | 0.087 | 0.646 |

*3.3. Accuracy Performance*

The evaluation of competing models' forecasts is one of the key factors in time series analysis. Five forecasting measures were used for accuracy evaluation in this study to test the validity and generality of several models for HIV cases in the Philippines. These performance factors used for accuracy analysis are NMRSE, MAE, MAPE, WAPE, and $R^2$. Results of 10 independent runs are shown in Table 4. In this table, five different performance metrics for accuracy were used to identify statistical differences between the two models. The table clearly demonstrates that the suggested approach consistently provides the best value under all possible conditions (min, max, and average), which points to a more accurate forecasting model.

The lowest NMRSE, MAE, MAPE, and WAPE are the best values, and therefore, they are being subjected to comparative analysis on the level of improvement of the proposed model. The model consistently provides the least errors over ANN-PSO in all types of conditions. On the other hand, the best value for $R^2$ is its highest or maximum value. This study produced the highest coefficient for determination, meaning it provides a better proportion of variance as against the old model.

A statistical test known as the t-test was employed to assess the averages of the two groups. With a 95% confidence level or 0.05 level of significance, it was used to assess if the suggested model had significantly improved the accuracy. To determine if the suggested approach, when tested against the ANN-PSO model, is statistically significant or not, a paired two-sample of all accuracy measures was performed using the t-Test Paired Two Sample for Means tool in Excel. The datasets represented the results of 10 experiments using various accuracy metrics. It assesses whether the suggested method has significant improvement in accuracy over ANN-PSO using the Paired t-test with 0.05 as the level of significance. Table 5 presents the t-test's findings.

The computed p-values for NRMSE (0.04889174), MAE (0.02829063), MAPE (0.02226053), WAPE (0.01701545), and $R^2$ (0.00000021) of the proposed algorithm are less than the set 0.05 level of significance; thus the values indicated a significant result in terms of improvement in accuracy performance (Table 5-a to Table 5-e).

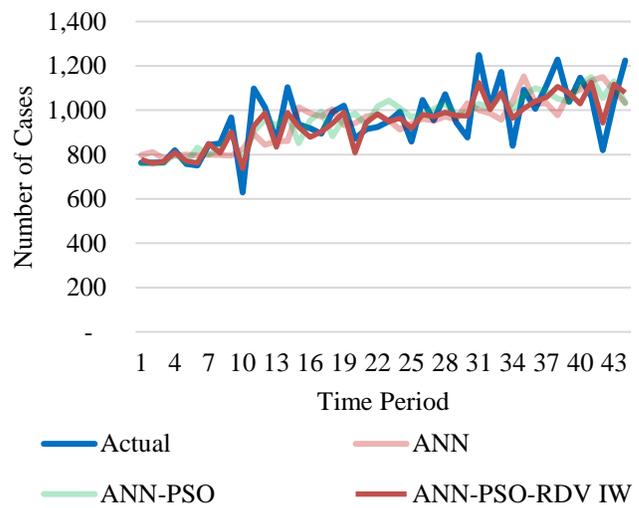

**Fig. 4 Comparison of actual vs forecast of the three models**

Table 5. Two-sample paired t-test of all accuracy metrics

| Algorithm | Mean | Std Dev | df | t-stat | p-value | Interpretation |
|---|---|---|---|---|---|---|
| **This Study** | 0.19577595 | 0.00301162 | 9 | -1.84721373 | 0.04889174 | significant |
| **ANN-PSO** | 0.19925778 | 0.00625766 | | | | |

(a) **NRMSE**

| Algorithm | Mean | Std Dev | df | t-stat | p-value | Interpretation |
|---|---|---|---|---|---|---|
| **This Study** | 81.09562808 | 1.36630209 | 9 | -2.18640021 | 0.02829063 | significant |
| **ANN-PSO** | 83.09120942 | 3.15594594 | | | | |

(b) **MAE**

| Algorithm | Mean | Std Dev | df | t-stat | p-value | Interpretation |
|---|---|---|---|---|---|---|
| **This Study** | 0.08683719 | 0.00190914 | 9 | -2.33309300 | 0.02226053 | significant |
| **ANN-PSO** | 0.08918308 | 0.00295275 | | | | |

(c) **MAPE**

| Algorithm | Mean | Std Dev | df | t-stat | p-value | Interpretation |
|---|---|---|---|---|---|---|
| This Study | 0.08513148 | 0.00157948 | 9 | -2.49696463 | 0.01701545 | significant |
| ANN-PSO | 0.08744769 | 0.00332141 | | | | |

(d) **WAPE**

| Algorithm | Mean | Std Dev | df | t-stat | p-value | Interpretation |
|---|---|---|---|---|---|---|
| **This Study** | 0.75963292 | 0.01969167 | 9 | 12.90181209 | 0.00000021 | significant |
| **ANN-PSO** | 0.64620540 | 0.01801480 | | | | |

(e) $R^2$

A comparison of actual and observed values of HIV cases for the three algorithms can be shown in Figure 4. The figure presents how close the actual values are to the forecast values produced by the proposed algorithm. Subsequently, the optimal solution point is achieved when ANN-PSO is integrated with the RDV IW technique.

## 4. Conclusion and Recommendations

The results of the study lead to the following conclusions. The optimal combination of RDV IW control parameters, which generates the least position error, is alpha = 0.4 and alpha_dump = 0.9. The ANN-PSO integrated with RDV IW provides better convergence performance compared to the old model, as shown by a 6.36% improvement in position error, 11.75% improvement in computational time, and 12.50% improvement in the number of iterations. Subsequently, the ANN-PSO integrated with RDV IW produced significant improvement in the accuracy performance, as shown by its computed p-value of NRMSE (0.04889174), MAE (0.02829063), MAPE (0.02226053), WAPE (0.01701545), and $R^2$ (0.00000021) which is less than the standard p-value of 0.05 in the significance level. Therefore, the integration of the RDV IW technique using an optimal combination [0.4, 0.9] provides better convergence and predictive accuracy to ANN-PSO in forecasting the number of HIV cases in the Philippines. On the basis of the study's findings, the future researcher may consider investigating the existing IW techniques such as random, random chaotic, and linear decreasing that enhance PSO and find out its convergence performance. Test the existing three forms of hybrid ANNs, such as TANN, CANN, and PANN, and find out their accuracy performance. Explore the use of extreme values of alpha_dump between [0.1, 0.45] to determine its convergence performance when paired with alpha values.